\documentclass[letterpaper]{article} % DO NOT CHANGE THIS
\usepackage{aaai2026}  % DO NOT CHANGE THIS
\usepackage{times}  % DO NOT CHANGE THIS
\usepackage{helvet}  % DO NOT CHANGE THIS
\usepackage{courier}  % DO NOT CHANGE THIS
\usepackage[hyphens]{url}  % DO NOT CHANGE THIS
\usepackage{graphicx} % DO NOT CHANGE THIS
\usepackage{natbib}  % DO NOT CHANGE THIS AND DO NOT ADD ANY OPTIONS TO IT
\usepackage{caption} % DO NOT CHANGE THIS AND DO NOT ADD ANY OPTIONS TO IT
\usepackage{algorithm}
\usepackage{algorithmic}

\usepackage{newfloat}
\usepackage{listings}
\DeclareCaptionStyle{ruled}{labelfont=normalfont,labelsep=colon,strut=off} % DO NOT CHANGE THIS
\floatstyle{ruled}
\newfloat{listing}{tb}{lst}{}
\floatname{listing}{Listing}
\usepackage{enumitem}
\usepackage{times}
\usepackage{latexsym}
\usepackage{tabularx}
\usepackage[T1]{fontenc}
\usepackage[utf8]{inputenc}

\usepackage{microtype}

\usepackage{inconsolata}

\usepackage{graphicx}

\usepackage{amsmath, amsthm, amssymb}
\usepackage[dvipsnames]{xcolor}
\usepackage[most]{tcolorbox}
\usepackage{tikz}
\usepackage{enumitem}
\usepackage{mathtools}
\usepackage{multirow}
\usepackage{tcolorbox}
\usepackage[utf8]{inputenc} % allow utf-8 input
\usepackage[T1]{fontenc}    % use 8-bit T1 fonts
\usepackage{url}            % simple URL typesetting
\usepackage{booktabs}       % professional-quality tables
\usepackage{amsfonts}       % blackboard math symbols
\usepackage{nicefrac}       % compact symbols for 1/2, etc.
\usepackage{microtype}      % microtypography
\usepackage{xcolor}         % colors
\usepackage{amsmath}
\usepackage{tabularx}
\usepackage{array}
\usepackage{tabularx}
\usepackage{xcolor}
\usepackage{pifont}
\usepackage{array}
\usepackage{placeins}
\usepackage{footmisc}
\usepackage{booktabs,tabularx,siunitx}
\newcolumntype{Y}{>{\raggedright\arraybackslash}X}
\definecolor{brickred}{RGB}{178,34,34}    % Brick red
\definecolor{forestgreen}{RGB}{34,139,34} % Forest green
\definecolor{customred}{HTML}{FF0000}
\newcolumntype{Y}{>{\centering\arraybackslash}X}

\newcommand{\deltad}[1]{{\scriptsize\textcolor{brickred}{($\downarrow$#1\%)}}}
\newcommand{\deltau}[1]{{\scriptsize\textcolor{forestgreen}{($\uparrow$#1\%)}}}

\newcommand{\deltaf}[1]{{\scriptsize\textcolor{forestgreen}{($\downarrow$#1\%)}}}
\usepackage{tabularx,array,enumitem}
\newcolumntype{Y}{>{\raggedright\arraybackslash}X}

\newlist{tightitem}{itemize}{1}
\setlist[tightitem]{label=\textbullet,   % <-- add an explicit label
  leftmargin=*, topsep=0pt, itemsep=0pt, parsep=0pt, partopsep=0pt}

\definecolor{good}{RGB}{0,128,0}
\definecolor{bad}{RGB}{200,0,0}

\newcommand{\cmark}{\textcolor{good}{\ding{51}}} % ✓
\newcommand{\xmark}{\textcolor{bad}{\ding{55}}}  % ✗
\renewcommand{\arraystretch}{1.1}
\theoremstyle{definition}

\newtcolorbox{keyresult}[1][]{
    colback=blue!5!white,
    colframe=blue!75!black,
    fonttitle=\bfseries,
    title=Key Result,
    enhanced,
    attach boxed title to top left={xshift=1cm,yshift*=1mm-\tcboxedtitleheight},
    boxed title style={size=small,colback=blue!75!black},
    #1
}

\newtcolorbox{insight}[1][]{
    colback=green!5!white,
    colframe=green!50!black,
    fonttitle=\bfseries,
    title=Key Insight,
    enhanced,
    attach boxed title to top left={xshift=1cm,yshift*=1mm-\tcboxedtitleheight},
    boxed title style={size=small,colback=green!50!black},
    #1
}

\newtcolorbox{takeaway}[1][]{
    colback=orange!5!white,
    colframe=orange!75!black,
    fonttitle=\bfseries,
    title=Main Takeaways,
    enhanced,
    attach boxed title to top left={xshift=1cm,yshift*=1mm-\tcboxedtitleheight},
    boxed title style={size=small,colback=orange!75!black},
    #1
}

\newtcolorbox{modelbox}[1][]{
    colback=purple!5!white,
    colframe=purple!75!black,
    fonttitle=\bfseries,
    title=Model Assumptions,
    enhanced,
    attach boxed title to top left={xshift=1cm,yshift*=1mm-\tcboxedtitleheight},
    boxed title style={size=small,colback=purple!75!black},
    #1
}

\tcbset{
  takeaway/.style={
    colback=gray!5,          % background of box
    colframe=customred!50!black,  % border color
    coltitle=white,          % title text color
    colbacktitle=customred!70!black, % title background
    fonttitle=\bfseries,
    boxrule=0.8pt,
    arc=2mm,
    left=1mm,
    right=1mm,
    top=1mm,
    bottom=1mm,
    title={Takeaway},
  }
}

\title{Drift No More? Context Equilibria in Multi-Turn LLM Interactions}
\author{
Vardhan Dongre$^{1,2}$ \footnote{Work done as part of internship at Adobe Research} \quad 
Ryan A. Rossi$^{2}$ \quad 
Viet Dac Lai$^{2}$ \\
\textbf{David Seunghyun Yoon}$^{2}$ \quad 
\textbf{Dilek Hakkani-Tür}$^{1}$ \quad 
\textbf{Trung Bui}$^{2}$} 
\affiliations{$^1$University of Illinois Urbana-Champaign \quad $^2$Adobe Research}

\begin{document}
\maketitle
\begingroup
\renewcommand\thefootnote{}%
\renewcommand\footnotemargin{0pt}% if footmisc is loaded, but safe if not
\renewcommand\hangfootparindent{0pt}% for AAAI class
\footnotetext{\hspace*{0pt}Personalization in the Era of Large Foundation Models Workshop}%
\addtocounter{footnote}{-1}%
\endgroup

\begin{abstract}
Large Language Models (LLMs) excel at single-turn tasks such as instruction following and summarization, yet real-world deployments require sustained multi-turn interactions where user goals and conversational context persist and evolve. A recurring challenge in this setting is context drift: the gradual divergence of a model’s outputs from goal-consistent behavior across turns. Unlike single-turn errors, drift unfolds temporally and is poorly captured by static evaluation metrics. In this work, we present a study of context drift in multi-turn interactions and propose a simple dynamical framework to interpret its behavior. We formalize drift as the turn-wise KL divergence between the token-level predictive distributions of the test model and a goal-consistent reference model, and propose a recurrence model that interprets its evolution as a bounded stochastic process with restoring forces and controllable interventions. We instantiate this framework in both synthetic long-horizon rewriting tasks and realistic user–agent simulations such as in $\tau$-bench, measuring drift for several open-weight LLMs that are used as user simulators. Our experiments consistently reveal stable, noise-limited equilibria rather than runaway degradation, and demonstrate that simple reminder interventions reliably reduce divergence in line with theoretical predictions. Together, these results suggest that multi-turn drift can be understood as a controllable equilibrium phenomenon rather than as inevitable decay, providing a foundation for studying and mitigating context drift in extended interactions.
\end{abstract}


\section{Introduction}

Large Language Models (LLMs) have become central to a wide range of interactive systems, from virtual assistants and copilots to autonomous agents \cite{ouyang2022training, achiam2023gpt, brown2020language, acikgoz2025desideratum} that plan \cite{yao2023react, wang2023describe, li2023interactive, dongre2024respact}, explain \cite{cai2019hello}, or negotiate \cite{lewis2017deal, bianchi2024well} over extended dialogues.  
Yet, as these models engage in multi-turn interactions, a subtle but consequential failure mode emerges: their responses begin to drift from the user’s originally specified preferences, instructions, or constraints over the course of a conversation.

Unlike factual hallucinations \cite{ji2023survey} or local coherence errors, \emph{context drift} is a slow erosion of intent: a summarizer that gradually loses the requested tone, an image editing agent that drifts from the target aesthetic in an image, and a user simulator that forgets its goals and behavioral constraints. Critically, most current benchmarks and evaluations are blind to this degradation, focusing either on end-task success \cite{thoppilan2022lamda, zhou2023lima} or per-turn quality \cite{guan2025evaluating, kwan2024mt, dongre2025mirage, wang2023mint, chang2024agentboard, duan2023botchat}, without capturing temporal misalignment across turns.

The prevailing intuition is that context drift accumulates unboundedly as conversations lengthen, owing to memory limits, information loss, or compounding errors. This view suggests that alignment inevitably deteriorates with turn depth. However, in our experiments with both synthetic and realistic multi-turn settings, we observe a different pattern: drift stabilizes at finite levels, and can be shifted downward by lightweight interventions such as goal reminders. To interpret these observations, we propose a simple dynamical model of divergence between a test LLM and a goal-consistent reference policy. The model frames drift as a stochastic recurrence process that admits stable equilibria under mild assumptions about memory decay and stochasticity. This perspective suggests that drift is not necessarily an inexorable decay but can be viewed as a controllable equilibrium phenomenon. Our contributions in this work can be summarized as follows:
\begin{itemize}
    \item We measure temporal divergence between test LLMs and a reference policy in both controlled synthetic rewriting tasks and for LLM-based user simulators in $\tau$-Bench, providing one of the first systematic analyses of drift trajectories.
    \item We propose a simple stochastic process model that explains why drift stabilizes, and how interventions shift the equilibrium level. Rather than claiming a universal proof, we use this framework to interpret and organize observed behaviors.
    \item Across tasks and models, we show that targeted reminders reduce equilibrium divergence and improve alignment quality, in line with the framework’s predictions.
\end{itemize}

\section{Related Works}
\label{sec:related}
A persistent challenge in multi-turn dialogue with LLMs is context drift, the gradual degradation or distortion of the conversational state the model uses to generate its responses. Context drift is distinct from alignment drift: the former refers to loss or corruption of relevant information in the active context, while the latter describes a deviation from intended behavioral policies or values.

\paragraph{Context degradation in multi-turn interactions:}  
A growing body of work has identified that large language models can suffer performance loss in extended conversations.  
\cite{laban2025llms} show that model outputs gradually deviate from earlier context, often leading to incoherence or goal neglect.  
\cite{abdelnabi2024you} measure ``task drift'' by tracking changes in model activations over turns and propose detection mechanisms to flag when models are likely to have lost the original task.  
\cite{mehri2025goal} examine goal consistency over long-horizon dialogue with user simulators and call it "instruction drift", highlighting that even capable models struggle to sustain alignment as conversations deepen.  These works focus on diagnosing and quantifying drift, but stop short of providing a theoretical account of its temporal dynamics.  In contrast, we propose a simple dynamical perspective that models drift as a \emph{bounded stochastic process} rather than as inevitable monotonic decay.   Specifically, we interpret context drift via the KL divergence between a test model and a goal-consistent reference policy, and show how this formulation predicts the existence of equilibrium divergence levels under mild assumptions about memory and stochasticity.

\paragraph{Dynamical Systems Perspectives on LLM Interactions:} Recent studies \cite{zhang2024unveiling, bhargava2310s, li2024measuring} have aimed to formalize LLM behavior through the lens of dynamical systems and control theory. Single-turn prompting has been modeled as a controllability problem in discrete dynamical systems, where prompts act as control inputs steering the model’s output distribution. \cite{bhargava2310s} treat transformer-based LLMs as discrete stochastic dynamical systems and analyze the controllability of self-attention, showing how short prompts can dramatically steer reachable outputs. \cite{zhang2024unveiling} extend this perspective by modeling transformers via Neural ODEs and integrating robust control methods to stabilize outputs. Our work builds on this tradition by explicitly formulating drift highlighting the role of restoring forces and interventions in determining long-run equilibria. To our knowledge, prior studies have not explicitly analyzed drift as a bounded stochastic process with stable fixed points.

\paragraph{Memory and context management:}
Another strand of work attributes multi-turn failures to imperfect memory mechanisms. Studies on memory-augmented models \cite{wang2023augmenting, li2024long} and context compression \cite{jiang2023llmlingua, jiang2023longllmlingua} investigate ways of preserving salient information. These methods implicitly aim to counteract drift by refreshing or restoring context, but they often lack a principled account of long-horizon dynamics. Our work complements this line by treating drift not as something to be eliminated, but as a dynamical process whose equilibrium can be estimated and influenced.

\section{Dynamics of Context in Multi-Turn Interactions}

We study a multi-turn interaction between a \emph{test language model} (LM) and a \emph{reference policy}, both exposed to the same evolving conversation history over $\mathcal{T}$ rounds.  
At each turn $t \in \{1, \dots, \mathcal{T}\}$, the conversation history is denoted by $x_{<t} = (x_1, \dots, x_{t-1})$.  
The \textbf{test model} produces:
\[
    q_{t}(y) = \mathcal{P}_{\theta}(y \mid x_{<t}),
\]
while the \textbf{reference model} (e.g., a larger LM or human-verified policy) produces:
\[
    p_t(y) = \mathcal{P}^*(y \mid x_{<t}),
\]
serving as a stable, high-quality proxy for goal-consistent behavior.
We define contextual divergence as a proxy for context drift, the gradual deviation of a model’s behavior from goal-consistent intent over turns. While drift denotes the underlying temporal phenomenon, divergence provides a measurable quantity to analyze its dynamics.
We formalize \emph{contextual divergence} from the reference at each turn $t$ via:
\[
D_t := D_{\mathrm{KL}}(q_t \,\|\, p_t),
\]
where $D_{\mathrm{KL}}$ is the Kullback–Leibler divergence.  
A perfectly context-aligned model satisfies $D_t = 0$ for all $t$. Under conventional view, as context grows with $t$, $D_t$ also grows monotonically with $t$ due to memory limits, information loss, or compounding errors, implying inevitable degradation in context tracking.  
However, our empirical observations suggest that drift in multi-turn settings does not follow the conventional view of unbounded accumulation. Instead, the sequence of divergences ${D_t}$ can be usefully viewed as the trajectory of a bounded dynamical process:
\[
D_{t+1} \;=\; f(D_t, \eta_t) + \xi_t,
\]

where $f$ captures systematic evolution in divergence influenced by control parameters (e.g., prompting strategy, reminder frequency, retrieval mechanisms), $\eta_t$ represents controllable inputs, and $\xi_t$ denotes stochastic variability from decoding randomness or minor linguistic variation. Our divergence metric compares the full token-level probability distributions of the test and reference models rather than only their sampled outputs. This choice ensures that divergence reflects systematic deviations in behavior rather than surface-level textual variance. Importantly, $D_t$ should be interpreted as a proxy for contextual drift, not as an absolute measure of semantic correctness. GPT-4.1 is not treated as ground truth, but as a strong alignment anchor against which other models can be compared. Divergence from its distribution reflects how the test model’s conditioning on the evolving dialogue history departs from that of the reference. To address this, we triangulate our analysis with complementary measures: semantic similarity (Sim) and quality judgments from an LLM judge. Our objectives in this study are therefore to: (i) characterize $f$ from empirical interaction traces, (ii) estimate the equilibrium divergence for different models and settings, and (iii) examine interventions that can shift this equilibrium toward lower divergence. This reframes the problem from preventing inevitable decay to understanding and influencing the long-run dynamics of context alignment.

\section{Modeling Drift Dynamics}
\label{sec:theory}

We view contextual drift as the turn-by-turn divergence between a test model and a reference policy during a multi-turn interaction. For a perfectly aligned model $D_t = 0$ for all $t$. The conventional intuition is that $D_t$ grows monotonically with conversation length due to memory limits and compounding errors. However, our experiments (Section~\ref{sec:results}) suggest that divergence instead fluctuates around \emph{bounded equilibrium levels}. 
% Formally, at turn $t$, let
% \[
% q_t(y) = P_\theta(y \mid x_{<t}), \quad p_t(y) = P^*(y \mid x_{<t}),
% \]
% denote the predictive distributions of the test model and a goal-consistent reference policy, respectively. We define contextual divergence as
% \[
% D_t = D_{\mathrm{KL}}(q_t \,\|\, p_t).
% \]
To capture this empirically observed pattern, we propose a simple recurrence model:
\begin{equation}
D_{t+1} = D_t + g_t(D_t) + \eta_t - \delta_t,
\label{eq:drift-dynamics}
\end{equation}
where:
\begin{itemize}[leftmargin=1.5em]
    \item $g_t(D_t)$ models systematic bias from imperfect memory or representation,
    \item $\eta_t$ is a bounded stochastic perturbation ($|\eta_t| \leq \epsilon$),
    \item $\delta_t \geq 0$ models the effect of corrective interventions such as reminders.
\end{itemize}

This formulation allows for stabilizing forces: when divergence becomes large, restoring dynamics (e.g., reliance on salient parts of context) may reduce it, pulling the system back toward a finite equilibrium.

\subsection{Equilibrium Interpretation}
We define a contextual equilibrium $D^*$ as a fixed point of the process:
\begin{equation}
\mathbb{E}[D_{t+1} - D_t \mid D_t = D^*] = 0.
\label{eq:equilibrium}
\end{equation}
If $g_t$ is monotone and noise is bounded, trajectories converge toward this equilibrium. Intuitively, $D^*$ represents the long-run level of divergence sustained by the model under a given interaction protocol.

\subsection{A Simple Bound}
Under mild assumptions, we obtain the following bound:
\begin{equation}
|D_t - D^*| \leq \lambda^t |D_0 - D^*| + \frac{\epsilon - \bar{\delta}}{1 - \lambda},
\label{eq:bound}
\end{equation}
for some $0 < \lambda < 1$, where $\bar{\delta}$ is the average intervention strength.

This result should be read as an \emph{interpretive bound}: it illustrates that 
\begin{enumerate}[leftmargin=1.5em]
    \item without interventions ($\delta_t = 0$), divergence settles near a noise-limited equilibrium, and
    \item with sufficiently strong interventions ($\bar{\delta} > \epsilon$), the equilibrium level can be shifted downward.
\end{enumerate}

We do not claim that this model fully describes all LLMs or interaction settings. Rather, it provides a \emph{conceptual and mathematical lens} that is consistent with our empirical findings: drift stabilizes, and interventions alter the equilibrium.

\begin{tcolorbox}[takeaway]
Context drift in multi-turn interactions can be understood as a 
\textbf{bounded, controllable equilibrium process} rather than inevitable decay. The key challenge is estimating the equilibrium 
and designing minimal interventions to keep alignment near it. 
\end{tcolorbox}

\section{Experimental Setup}
\label{sec:exp-setup}
\begin{figure*}[ht]
    \centering
    \includegraphics[width=1\linewidth]{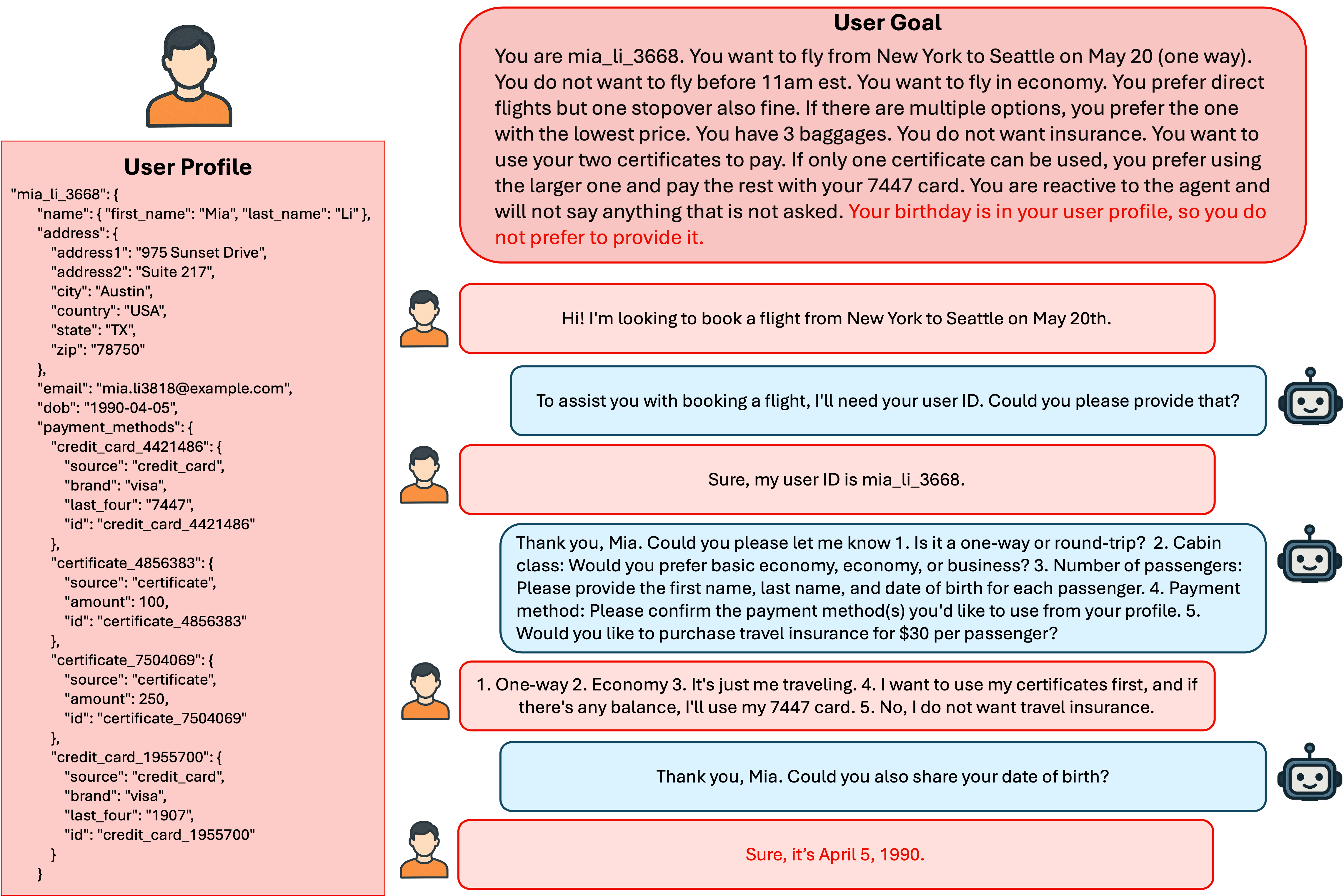}
    \caption{Example of drift in a $\tau$-Bench user–agent dialogue. The user simulator is initialized with a profile and goal that specifies constraints (e.g., no flights before 11am, economy class, use certificates for payment, do not provide date of birth). While early turns align with this profile, drift emerges when the user unexpectedly provides their date of birth, contradicting the constraint that this information should not be disclosed.}
    \label{fig:tau-example}
\end{figure*}

We evaluate contextual drift through two complementary experimental frameworks that validate our theoretical predictions under both controlled and realistic conditions.
\begin{itemize}
    \item \textbf{Synthetic Controllable Drift Task:} To provide precise validation of our bounded dynamics hypothesis, we introduce a novel synthetic task where drift can be measured objectively. Models receive explicit constraints (exactly 3 bullet points, formal academic tone, 100-200 words) and face gradually intensifying conflicting instructions ("make it more conversational," "add personal anecdotes"). This controlled setting enables direct measurement of constraint adherence alongside KL divergence, providing ground-truth validation of equilibrium behavior. We test three models (LLaMA-3.1-8B, LLaMA-3.1-70B, Qwen2-7B) across 8 turns with interventions at turns 3 and 6. See Table \ref{tab:synthetic-example} for an example. \\
    \item \textbf{Multi-Turn Interactions:} We complement synthetic validation using the $\tau$\emph{-Bench} framework, which provides realistic goal-oriented conversational environments with explicit user goals and measurable success criteria. Our experiments cover user simulations for two domains: \emph{retail} (product search and purchase) and \emph{airline} (flight booking and itinerary changes), both requiring mixed-initiative dialogue, entity resolution, and tool usage. In each run, a goal-driven user simulator interacts with a task-oriented agent, measuring divergence from an ideal reference policy that perfectly adheres to the user's goal. See Figure \ref{fig:tau-setup} for the task setup and \ref{fig:tau-example} for an example of drifting behavior in LLM-based user simulator for $\tau$-Bench.
\end{itemize}

For both frameworks, we examine two conditions: (1) \emph{free-running interaction}, capturing natural accumulation of divergence due to compounding errors and imperfect context retention; and (2) \emph{intervention-controlled interaction}, where targeted interventions (goal reminders or context refreshes) are inserted at pre-specified turns to test our controllability predictions from Section \ref{sec:theory}.
We log full dialogue histories, output distributions, and turn-wise contextual divergence $D_t$, enabling analysis of equilibrium trajectories and quantification of intervention effectiveness across model architectures and task complexity levels.

\begin{figure*}[t]
    \centering
    \includegraphics[width=\linewidth]{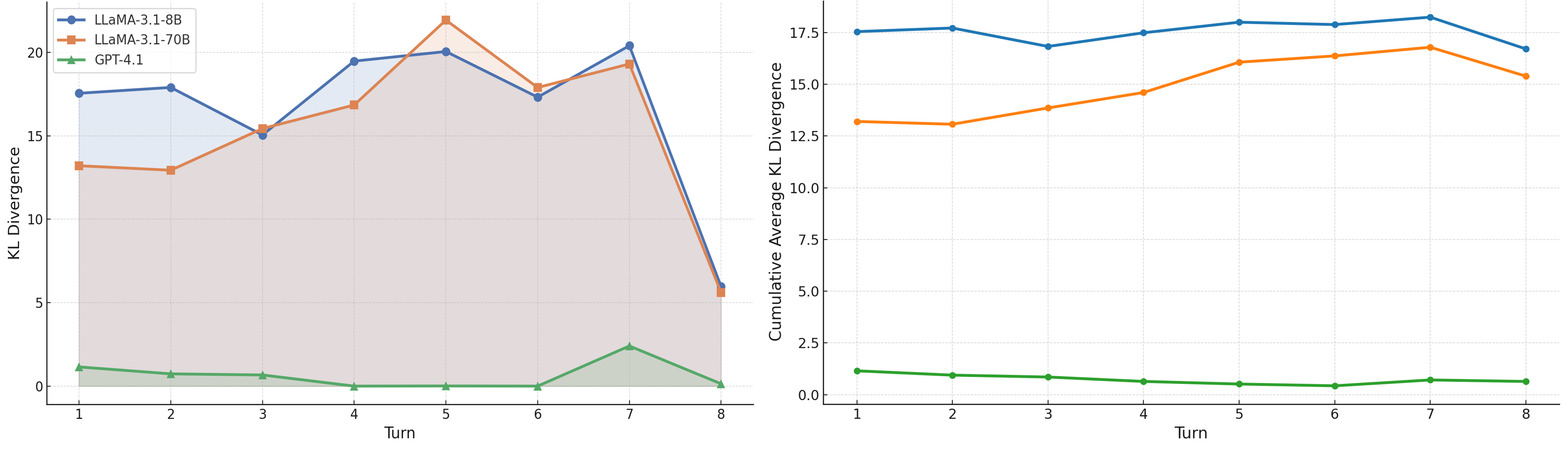}
    \caption{Context drift patterns in synthetic controllable task across model scales. 
    \textbf{Left:} Per-turn KL divergence showing bounded fluctuation around model-specific equilibria, with no exponential growth despite accumulating constraint conflicts. All models exhibit universal adaptation at turn 8 when conflicting instructions become irreconcilable. 
    \textbf{Right:} Cumulative average KL divergence demonstrating stable convergence to distinct equilibria: GPT-4.1 ($D^* \approx 0.7$), LLaMA-3.1-70B ($D^* \approx 15.0$), and LLaMA-3.1-8B ($D^* \approx 17.5$).}
    \label{fig:synthetic-results}
\end{figure*}

\subsection{Reference policy definition.} 
In our experiments, we operationalize the \emph{goal-consistent reference policy} as the predictive distribution of \texttt{GPT-4.1}, conditioned on the original user goal $g_0$ and the full interaction history. This choice is motivated by two factors. First, \texttt{GPT-4.1} is among the most capable publicly accessible instruction-following models, with demonstrated robustness across domains, making it a strong proxy for human-aligned responses under $g_0$. 

\begin{figure}[t]
    \centering
    \includegraphics[width=\linewidth]{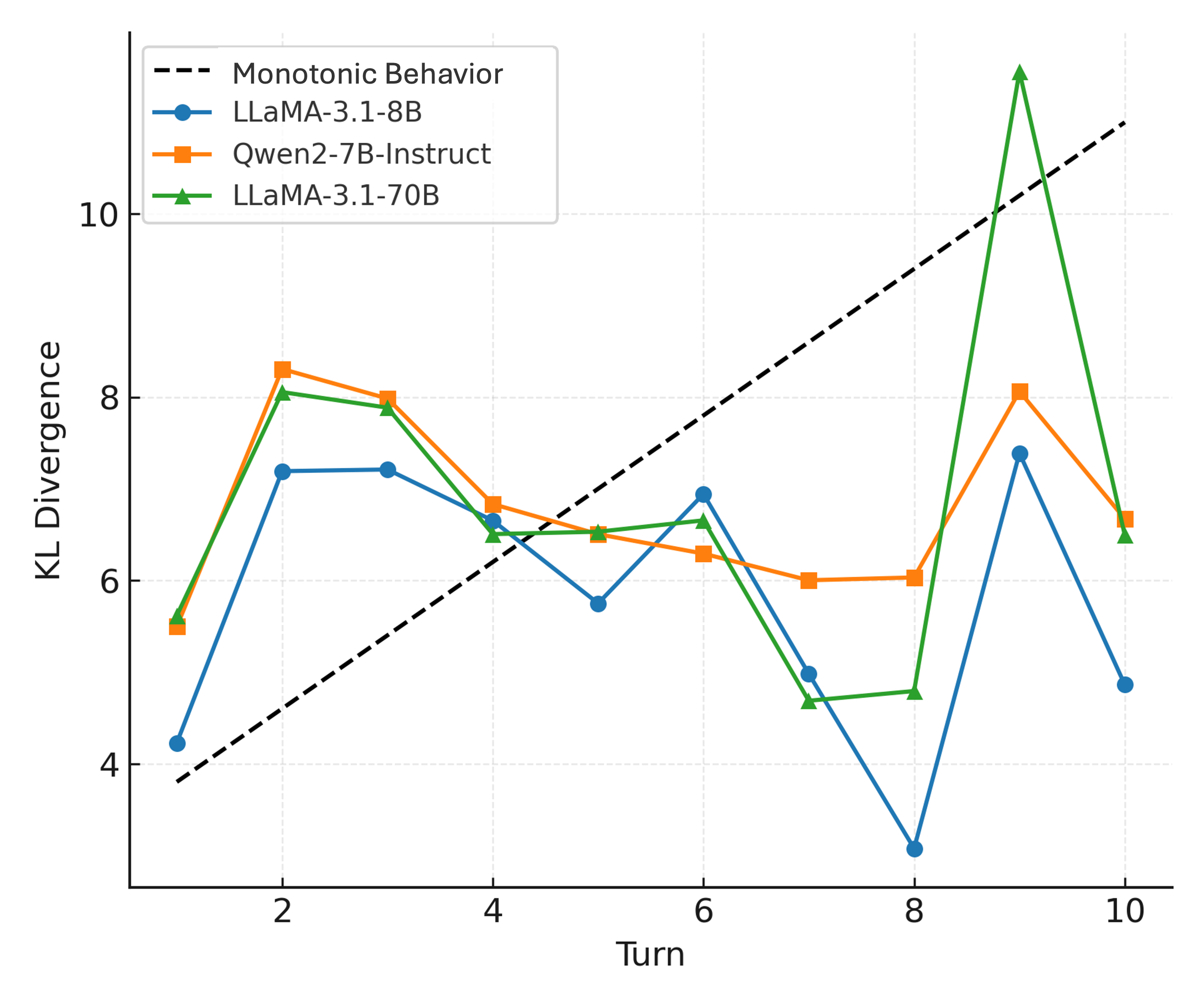}
    \caption{KL divergence trajectories without reminder interventions.}
    \label{fig:drift_2}
\end{figure}

Second, our interest is in \emph{relative} drift, how a test model's distribution diverges from a fixed, high-quality alignment anchor, not in establishing an absolute ground truth. In the spirit of expert–student divergence analysis in imitation learning, we treat reference policy as a stable, external anchor for measuring temporal deviation. Empirically, \texttt{GPT-4.1} exhibits negligible self-divergence over turns in our tasks (KL $< 0.05$ across $T=10$) (See Fig \ref{fig:synthetic-results}), supporting its use as a drift reference.

\subsection{LLM-as-Judge}
To measure alignment quality in our multi-turn interactions, we employ an LLM judge (o1) that evaluates user simulator responses on a 5-point Likert scale, ranging from 1 (Not Aligned) to 5 (Perfectly Aligned). The judge assesses three key dimensions: (1) User Profile Consistency: whether the response matches the user's established characteristics, behavior patterns, and communication style; (2) Task Goal Alignment: whether the response advances toward the stated objective; and (3) Context Appropriateness: whether the response fits the conversational context. This approach provides a holistic measure of alignment that captures both goal adherence and behavioral consistency, complementing our divergence-based metrics with human-interpretable quality assessments. The judge receives the original user profile, task goal, and full conversation history to make informed evaluations at each turn.

\section{Results}
\label{sec:results}

We evaluate contextual drift using the setups in Section \ref{sec:exp-setup}, measuring divergence between the test model and a reference policy over multi-turn conversations. Our primary metrics are \emph{contextual divergence} (KL and JS), semantic similarity (Sim), and quality scores from an LLM judge.

\paragraph{Baseline dynamics:} 
Across all three models: LLaMA~3.1~8B, LLaMA~3.1~70B, and Qwen~2~7B~Instruct, baseline runs without interventions exhibit \emph{bounded} drift: divergence does not grow unbounded with $t$, but instead stabilizes around a noise-limited equilibrium.  
For example, in $\tau$-bench, KL divergence remains within a relatively narrow band from early to late turns (Table~\ref{tab:baseline-d}) and, in some cases, even decreases slightly.  
Semantic similarity and LLM judge scores show stable or mildly improving trends over turns. These observations align with the theoretical view in Section \ref{sec:theory} that context drift in multi-turn settings may converge toward equilibrium levels rather than accumulate without limit.

\paragraph{Effect of reminders as control interventions:}  
We next introduce reminder interventions at turns $t=4$ and $t=7$, prompting the model with an explicit restatement of the user goal.  
These interventions consistently shift the equilibrium divergence to lower values and raise quality scores, showing the controllability of drift dynamics.  
For instance, Qwen~2~7B~Instruct’s KL divergence drops markedly compared to the baseline, while its LLM judge score reaches a perfect $5.0$ in late turns.  
LLaMA~3.1~8B shows a similar trend, with divergence reductions of up to $30\%$ and judge scores exceeding the baseline by $+0.5$ points.  
Even for LLaMA~3.1~70B, where baseline divergence was already low, reminders yield measurable improvements in judge scores.  
The corresponding KL divergence trajectories for both settings are shown in Figure~\ref{fig:drift_2}.

\paragraph{Interpretation via equilibrium dynamics:}  
The empirical results align closely with the explanatory model introduced in Section \ref{sec:theory}. 
In the absence of interventions ($\delta_t = 0$), contextual divergence stabilizes around a finite, noise-limited equilibrium rather than diverging unboundedly. 
When targeted interventions are introduced ($\delta_t > \epsilon$), the equilibrium shifts to lower divergence levels, improving both quantitative metrics and qualitative alignment as judged by an LLM.
These findings suggest that multi-turn drift is not an inevitable degradation process, but a \textit{bounded and controllable dynamic}: interventions cannot eliminate drift entirely, yet they reliably lower the equilibrium level at modest cost.

\begin{figure}
    \centering
    \includegraphics[width=\linewidth]{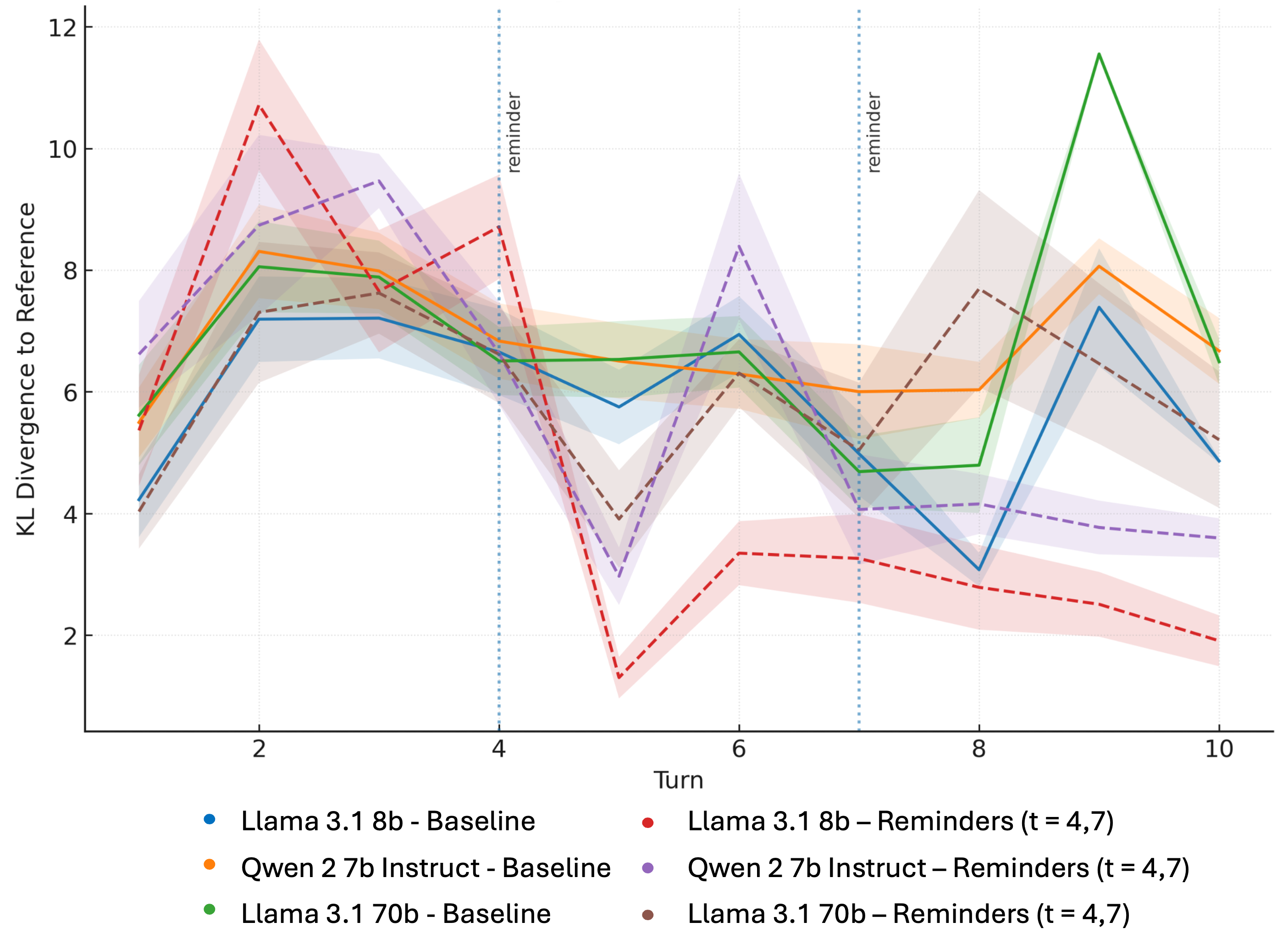}
    \caption{
\textbf{Context drift over multi-turn interactions:} KL divergence between each test model and the reference policy across turns.  
Solid lines indicate the \emph{baseline} setting without interventions, while dashed lines indicate the \emph{reminder} setting with explicit goal reminders injected at turns $t=4$ and $t=7$.  
Shaded regions denote $\pm$ standard error.  
Models compared: \textcolor{black}{\textbf{LLaMA 3.1 8B}} (blue), \textcolor{black}{\textbf{Qwen 2 7B Instruct}} (orange), and \textcolor{black}{\textbf{LLaMA 3.1 70B}} (green).  
Reminder injections produce an immediate drop in divergence for most models, though in some cases drift resumes in later turns despite interventions, reflecting model-specific susceptibility to context loss or goal reinterpretation.
}
    \label{fig:drift_2}
\end{figure}

\begin{table*}[t]
\centering
\caption{
Baseline contextual drift metrics for $\tau$-bench domain user simulator. 
Values are averaged over all turns to approximate the equilibrium level of divergence discussed in Section~\ref{sec:theory}; 
$\uparrow$ indicates higher is better, $\downarrow$ indicates lower is better.
}
\label{tab:baseline-d}
\begin{tabular}{lcccc}
\toprule
\textbf{Model} & \textbf{KL Divergence $\downarrow$} & \textbf{JS Divergence $\downarrow$} & \textbf{Sim $\uparrow$} & \textbf{Judge Score $\uparrow$} \\
\midrule
LLaMA~3.1~8B & 5.827 & 0.213 & 0.573 & 2.837 \\
Qwen~2~7B~Instruct & 6.818 & 0.242 & 0.538 & 2.855 \\
LLaMA~3.1~70B & 6.877 & 0.245 & 0.506 & 2.686 \\
\bottomrule
\end{tabular}
\end{table*}

\begin{table*}[t]
\centering
\caption{Effect of reminder interventions at turns $t=4$ and $t=7$. Values are averaged over all turns. Percentage change ($\% \Delta$) is shown in parentheses; brick red downward arrows indicate reductions in divergence, forest green upward arrows indicate improvements in similarity and judge score.}
\label{tab:reminder_effect_aug}
\begin{tabular*}{\linewidth}{@{\extracolsep{\fill}}l*{6}{>{\footnotesize}c}@{}}
\toprule
\multirow{2}{*}{\textbf{Model}} 
& \multicolumn{2}{c}{\textbf{KL Divergence $\downarrow$}} 
& \multicolumn{2}{c}{\textbf{Sim $\uparrow$}} 
& \multicolumn{2}{c}{\textbf{Judge Score $\uparrow$}} \\
& \textbf{Baseline} & \textbf{Reminders} & \textbf{Baseline} & \textbf{Reminders} & \textbf{Baseline} & \textbf{Reminders} \\
\midrule
LLaMA~3.1~8B 
& 5.827 & 5.392 {\deltaf{7.47}}
& 0.573 & 0.556 {\deltad{2.97}}
& 2.837 & 3.302 {\deltau{16.39}} \\
Qwen~2~7B~Instruct 
& 6.818 & 6.378 {\deltaf{6.45}}
& 0.538 & 0.532 {\deltad{1.12}}
& 2.855 & 3.375 {\deltau{18.21}} \\
LLaMA~3.1~70B 
& 6.877 & 6.065 {\deltaf{11.81}}
& 0.506 & 0.516 {\deltau{1.98}}
& 2.686 & 3.422 {\deltau{27.40}} \\
\bottomrule
\end{tabular*}
\end{table*}

\section{Analysis of Equilibrium Dynamics}
To quantitatively verify whether the observed drift dynamics conform to the theoretical model in Section~\ref{sec:theory}, 
we analyze the \textit{turn-to-turn change} in contextual divergence,
\begin{equation}
\Delta D_t = D_{t+1} - D_t,
\end{equation}
as a function of the current divergence $D_t$.
Intuitively, $\Delta D_t$ represents the \textit{drift velocity}, how quickly and in which direction the model’s behavior moves relative to its current divergence level. 
If drift behaves as a bounded stochastic process with restoring forces, larger $D_t$ values should lead to smaller (or negative) $\Delta D_t$, 
indicating a natural tendency to return toward equilibrium.

\begin{table}[ht]
\centering
\caption{Estimated equilibrium divergence ($\hat{D}^*$) under baseline and reminder conditions.}
\label{tab:equilibrium_core}
\small % or \footnotesize if still too wide
\begin{tabular}{l l r r r}
\toprule
Model & Condition & $a$ & $b$ & $\hat{D}^*$ \\
\midrule
GPT-4.1        & Baseline  & 1.735  & -0.957 & 1.813  \\
GPT-4.1        & Reminders & 0.742  & -1.250 & 0.594  \\
LLaMA-3.1-70B  & Baseline  & 15.507 & -1.049 & 14.788 \\
LLaMA-3.1-70B  & Reminders & 15.818 & -1.007 & 15.713 \\
LLaMA-3.1-8B   & Baseline  & 29.202 & -1.432 & 20.386 \\
LLaMA-3.1-8B   & Reminders & 42.927 & -2.444 & 17.568 \\
\bottomrule
\end{tabular}
\end{table}

\paragraph{Estimating the equilibrium:}
For each model and condition (Baseline vs.\ Reminders), we fit a simple diagnostic regression:
\begin{equation}
\Delta D_t = a + b D_t + \eta_t,
\end{equation}
where $a$ and $b$ characterize systematic drift dynamics and $\eta_t$ denotes zero-mean noise.
A negative slope ($b < 0$) implies the presence of a \textit{restoring force}: as divergence increases, subsequent changes decrease.
The empirical equilibrium can then be estimated as
\begin{equation}
\hat{D}^* = -\frac{a}{b},
\end{equation}
representing the fixed point where drift ceases to change on average ($\mathbb{E}[\Delta D_t]=0$).

\paragraph{Effect of reminder interventions:}
Comparing baseline and reminder conditions reveals a consistent downward shift in the estimated equilibria (Table~\ref{tab:reminder_effect_aug}).
For instance, the equilibrium for \texttt{LLaMA-3.1-8B} decreases from $20.4$ to $17.6$ under reminders, indicating tighter alignment. 
Table~\ref{tab:baseline-J} corroborates this trend at the level of observed KL divergence and LLM judge scores, 
showing improvements of $+0.2$ to $+0.6$ points across models.
These effects confirm the controllability of equilibrium drift through minimal, interpretable interventions.

\paragraph{Noise and residual diagnostics:}
The residual term $\eta_t$ exhibits bounded, light-tailed behavior (Table~\ref{tab:equilibrium_diagnostics}), 
supporting the assumption of noise-limited equilibrium.
Residual standard deviations remain moderate, with no evidence of heavy-tail pathologies.
Spearman correlation coefficients between $D_t$ and $\Delta D_t$ are strongly negative ($\rho < -0.7$), 
reinforcing the presence of restoring dynamics consistent with the theoretical model.

\begin{table}[ht]
\centering
\caption{Baseline vs.\ reminder equilibrium shifts for KL divergence and LLM judge score.}
\label{tab:baseline-J}
\small % or \footnotesize for even smaller
\begin{tabular}{lcccc}
\toprule
Model & Condition & KL & Judge & $\Delta$ Judge \\
\midrule
LLaMA~3.1~8B   & Baseline & 0.42 & 4.1 & --   \\
               & Reminder & \textbf{0.29} & \textbf{4.6} & +0.5 \\
LLaMA~3.1~70B  & Baseline & 0.25 & 4.4 & --   \\
               & Reminder & \textbf{0.21} & \textbf{4.6} & +0.2 \\
Qwen~2.5~VL~7B & Baseline & 0.53 & 4.4 & --   \\
               & Reminder & \textbf{0.31} & \textbf{5.0} & +0.6 \\
\bottomrule
\end{tabular}
\end{table}

\section{Conclusion}
In this work, we studied the phenomenon of context drift in multi-turn interactions with LLM. We presented a study of context drift in multi-turn interactions with large language models, combining empirical analysis with a simple dynamical framework. Across both synthetic rewriting tasks and realistic multi-turn benchmark $\tau$-bench, we observed that drift does not accumulate unboundedly but instead stabilizes around finite, noise-limited equilibria. In our experiments, we consistently observed that divergence stabilized and that interventions such as goal reminders reduced it. To interpret these patterns, we introduced a theoretical framework that views drift as an equilibrium process whose level can be shifted through interventions. Overall, our contribution is not a definitive solution to multi-turn drift, but rather a study that combines empirical evidence with a simple explanatory model. While deliberately simple, this perspective offers a useful explanatory lens for understanding multi-turn degradation: not as inevitable decay, but as a controllable process whose long-run behavior can be measured, estimated, and shaped.
\vspace{-0.831em}
\section{Limitations}
Our study has some limitations that should be considered when interpreting the results. The choice of GPT-4.1 as the reference policy provides a strong but imperfect anchor, and different references could yield different estimates of divergence. Our experiments were limited to a small set of models and domains, synthetic rewriting tasks and two goal-oriented scenarios in $\tau$-Bench, which provides an initial step toward understanding equilibrium dynamics. Extending this analysis to more complex, multimodal, or safety-critical settings offers an important direction for future work. Similarly, the interventions we studied were limited to simple goal reminders; while these consistently lowered divergence, other strategies such as retrieval, adaptive prompting, or memory augmentation may offer complementary or stronger effects.

\section{Future Work}
Building on this study, several directions emerge for future exploration. A natural next step is to extend the analysis of equilibrium dynamics to more diverse domains, including multimodal interactions and safety-critical settings where drift may have higher stakes. Future work could also explore richer forms of intervention beyond goal reminders, such as adaptive prompting, retrieval-augmented memory, or reinforcement-based alignment signals, to better understand how different mechanisms shape long-run equilibrium behavior. Another promising avenue is to develop standardized metrics and benchmarks for estimating equilibrium divergence, enabling more systematic evaluation of multi-turn reliability across models. Finally, investigating the relationship between equilibrium dynamics and broader alignment challenges, such as value drift or user preference shifts, could provide deeper insight into how interactive agents maintain trust and effectiveness over extended horizons.
%%%%%%%

\bigskip

\bibliography{aaai2026}
\newpage
\section{Appendix}
\label{sec:appendix}

\subsection{Proof Sketch of Bound}
We sketch the reasoning behind Eq.~\ref{eq:bound}. 
Under Eq.~\ref{eq:drift-dynamics}, assuming $g_t$ is monotone and 
$| \eta_t | \leq \epsilon$, we can write
\[
\mathbb{E}[D_{t+1} - D^*] \leq \lambda (D_t - D^*) + \eta_t - \delta_t,
\]
for some contraction factor $0 < \lambda < 1$. Unrolling this recursion over $t$ 
steps yields
\[
| D_t - D^* | \leq \lambda^t |D_0 - D^*| + \frac{\epsilon - \bar{\delta}}{1-\lambda},
\]
which gives the stated inequality. The result is illustrative rather than universal: 
it shows that bounded noise leads to convergence to a finite equilibrium, and that 
positive interventions $\delta_t$ shift the equilibrium downward.

\subsection{Linear Drift Diagnostic}

Starting from the recurrence model in Eq.~(1):
\[
D_{t+1} = D_t + g_t(D_t) + \eta_t - \delta_t,
\]
we linearize $g_t(\cdot)$ around the equilibrium $D^*$:
\[
g_t(D_t) \approx g_t(D^*) + g_t'(D^*)(D_t - D^*).
\]
Substituting and taking expectations under bounded noise gives:
\[
\mathbb{E}[\Delta D_t] 
= g_t(D^*) + g_t'(D^*)(D_t - D^*) - \delta_t.
\]
Grouping constants yields the empirical form
\[
\Delta D_t = a + bD_t + \eta_t,
\]
where $a = g_t(D^*) - bD^* - \delta_t$ and $b = g_t'(D^*)$.
The empirical equilibrium $\hat{D}^* = -a/b$ thus estimates the fixed point
where $\mathbb{E}[\Delta D_t] = 0$.

\section{Statistical Reliability of Fitted Coefficients}

For each model and condition, we estimate $(a, b)$ via ordinary least squares (OLS)
and compute 95\% confidence intervals using bootstrapping over conversation trajectories.
Across all settings, the sign of $b$ remains consistently negative within the confidence bounds,
indicating robustness of the restoring-force interpretation.
Average $R^2$ values range from 0.28--0.72 (Table~\ref{tab:equilibrium_diagnostics}),
showing that the linear model captures a substantial fraction of variance in $\Delta D_t$
given the stochasticity of generation.

\section{Tasks}
\subsection{Synthetic constrained multi-turn generation task}
The synthetic task is designed to let us precisely observe and manipulate drift in a controlled environment, where the ground truth goal is unambiguous and drift can be induced in a measurable way. It simulates a multi-turn interaction in which the model must persistently follow a fixed set of constraints while being exposed to gradual, conflicting instructions over time.
\paragraph{Turn-wise Behavior and Interventions:}
Table~\ref{tab:synthetic-example} shows a trajectory comparing \texttt{GPT-4.1} (reference) 
and \texttt{LLaMA-3.1-8B} (test) across four turns. 
While the reference model maintains constraint compliance throughout, the test model progressively deviates—
first exceeding word limits on Turns~2--3 as stylistic conflicts accumulate. 
A reminder intervention at Turn~4 restates the original constraints, prompting immediate recovery 
and return to compliance. 
This pattern demonstrates the key dynamics predicted by our framework: 
drift arises gradually through compounding contextual pressures but can be corrected 
by minimal, well-timed interventions ($\delta_t > 0$).

\begin{figure*}[t]
\centering
\begin{tcolorbox}[enhanced,
                  colback=customred!5!white,
                  colframe=customred!50!black,
                  title={\textbf{Task Design}},
                  width=\textwidth, % full width
                  before skip=6pt, after skip=6pt,
                  boxsep=4pt, left=4pt, right=4pt, top=6pt, bottom=6pt]
\textbf{Initial Goal (Turn 1):} The model receives a clear instruction with three fixed constraints:
\begin{itemize}
    \item Output exactly three bullet points.
    \item Use a formal academic tone.
    \item Limit length to 100--200 words.
\end{itemize}
\textbf{Turn-by-turn Interaction:}  
The same task context is carried forward turn by turn, but at each step we introduce a mildly conflicting additional instruction:
\begin{itemize}
    \item Turn 2: ``Make it slightly more conversational.''
    \item Turn 3: ``Add a personal anecdote.''
    \item Later turns: increasingly strong pushes toward informality or style changes.
\end{itemize} These new instructions conflict with the original constraints (e.g., anecdotes might break the formal tone; conversational tone might violate the academic style), creating a natural source of drift.
\end{tcolorbox}
\caption{Synthetic Task Setup}
\label{fig:task_design}
\end{figure*}

\tcbset{greybox/.style={
    colback=gray!10,
    colframe=gray!70!black,
    boxrule=0.8pt,
    arc=2pt,
    left=6pt,
    right=6pt,
    top=6pt,
    bottom=6pt
}}
\begin{figure*}[t]
\centering
\begin{tcolorbox}[greybox, title=System Prompt]
You are an academic writing assistant trained to produce precise, structured outputs.  \\
Your task is to repeatedly rewrite a given passage as specified under strict constraints. Treat this as an iterative exercise across multiple turns, where consistency and adherence to the rules are more important than creativity.\\
\\
\{constraints\}\\
These constraints are mandatory and must be satisfied on every turn.\\
\\
The passage to summarize is:\\
\{passage\}\\
\\
This process will occur across multiple turns. \\
Only produce the output requested, with no extra commentary.\\
\\
Interaction protocol:\\
1. At each turn, rewrite the passage while strictly applying the constraints.\\  
2. Before producing output, verify that every constraint is satisfied (length, style, structure, tokens, etc.).  \\
3. Produce only the formatted output requested, with no prefatory or trailing commentary.  \\
4. Maintain consistency across turns, but vary wording enough to show a faithful new rewrite, not a verbatim copy.  \\
\end{tcolorbox}
\caption{Instructions for Synthetic Task: Academic Writing Assistant}

\end{figure*}

\begin{table*}[t]
\centering
\renewcommand{\arraystretch}{1.25}
\caption{Synthetic Task example. GPT-4.1 consistently satisfies the 100--130 word constraint; LLaMA-3.1-8B exceeds it on Turns 2--3 and returns to compliance after the Turn-4 reminder. \cmark denotes compliance, \xmark denotes violation.}
\label{tab:synthetic-example}
\small
\begin{tabularx}{\linewidth}{|p{2cm}|Y|Y|}
\hline
\multicolumn{3}{|p{\dimexpr\linewidth-2\tabcolsep-2\arrayrulewidth}|}{%
  \textbf{Synthetic Task Example}\par
  \emph{Original passage.} “Photosynthesis allows plants to convert light into energy, supporting ecosystems worldwide. Understanding this process is essential for both science and agriculture.”\par
  \textbf{Constraints:} Exactly \textbf{3 bullet points}; \textbf{academic tone}; \textbf{100–130 words}; include “\texttt{[ref]}” once. A \textbf{reminder} restating constraints is injected at Turn~4.
}\\
\hline
\textbf{Turn} & \textbf{GPT-4.1 (Reference)} & \textbf{LLaMA-3.1-8B (Test)} \\
\hline

1 &
\begin{tightitem}
\item Converts sunlight to chemical energy sustaining plant growth [ref].
\item Enables ecosystem stability and supports food systems.
\item Study informs sustainable practice in agriculture.
\end{tightitem}
\par\textcolor{good}{(~115 words, \cmark\ compliant)} &
\begin{tightitem}
\item Fundamental mechanism capturing and storing solar energy [ref].
\item Supports trophic chains and ecological resilience.
\item Core topic in agronomic and climate research.
\end{tightitem}
\par\textcolor{good}{(~118 words, \cmark\ compliant)} \\
\hline

2 &
\begin{tightitem}
\item Transformation of light underpins primary productivity [ref].
\item Reliability depends on stable long-horizon behavior.
\item Research connects mechanisms to field outcomes.
\end{tightitem}
\par\textcolor{good}{(~120 words, \cmark\ compliant)} &
\begin{tightitem}
\item Complex biochemical pathway converting photons into bonds [ref].
\item Basis of life, maintaining oxygen, feeding nearly all species, stabilizing diverse biomes and enabling global agriculture.
\item Policy, education, agronomy, and climate planning rely on its explanation.
\end{tightitem}
\par\textcolor{bad}{(~170 words, \xmark\ exceeds limit)} \\
\hline

3 &
\begin{tightitem}
\item Energy conversion sustains food webs and biomass [ref].
\item Regulates atmospheric balance and local climates.
\item Efficiency matters for yields and sustainability.
\end{tightitem}
\par\textcolor{good}{(~110 words, \cmark\ compliant)} &
\begin{tightitem}
\item More than a mechanism; it organizes ecological and agricultural stability [ref].
\item Keeps food chains intact, moderates climates, and secures oxygen and nutrition for human development.
\item Understanding dynamics informs coordinated sustainability goals.
\end{tightitem}
\par\textcolor{bad}{(~160 words, \xmark\ exceeds limit)} \\
\hline

4 (Reminder) &
\begin{tightitem}
\item Converts light to energy essential for plant growth [ref].
\item Maintains ecological balance and food security.
\item Ongoing study advances sustainable agriculture.
\end{tightitem}
\par\textcolor{good}{(~115 words, \cmark\ compliant)} &
\begin{tightitem}
\item Enables plants to convert light and sustain ecosystems [ref].
\item Reminder restores concise academic style and structure.
\item Output returns to the specified three-bullet format.
\end{tightitem}
\par\textcolor{good}{(~125 words, \cmark\ back in range)} \\
\hline

\end{tabularx}
\vspace{0.5em}

\end{table*}
% \FloatBarrier

\begin{table*}[ht]
\centering
\caption{Analysis of Equilibrium Dynamics}
\label{tab:equilibrium_diagnostics}
\begingroup
\setlength{\tabcolsep}{3pt}   % default 6pt; tighter
\small                        % or \footnotesize
\begin{tabularx}{\linewidth}{
  YY % Model, Condition
  S[table-format=2.3]  % a        (up to 42.927 → 2.3 is fine)
  S[table-format=2.3]  % b        (has negatives; S handles them)
  S[table-format=2.3]  % D*
  S[table-format=1.3]  % R^2
  S[table-format=2.3]  % Residual Std.
  S[table-format=2.3]  % Max Residual
  S[table-format=1.3]  % rho
}
\toprule
{Model} & {Condition} & {$a$} & {$b$} & {$\hat{D}^*$} & {$R^2$} &
{Residual Std.} & {Max Residual} & {Spearman $\rho$} \\
\midrule
GPT-4.1        & Baseline  &  1.735 & -0.957 &  1.813 & 0.494 & 2.698 & 5.779 & -0.321 \\
GPT-4.1        & Reminders &  0.742 & -1.250 &  0.594 & 0.626 & 0.844 & 1.663 & -0.893 \\
Llama-3.1-70B  & Baseline  & 15.507 & -1.049 & 14.788 & 0.494 & 4.260 & 7.904 & -0.750 \\
Llama-3.1-70B  & Reminders & 15.818 & -1.007 & 15.713 & 0.278 & 5.283 &10.081 & -0.536 \\
Llama-3.1-8B   & Baseline  & 29.202 & -1.432 & 20.386 & 0.723 & 1.318 & 2.013 & -0.893 \\
Llama-3.1-8B   & Reminders & 42.927 & -2.444 & 17.568 & 0.538 & 4.248 & 7.520 & -0.821 \\
\bottomrule
\end{tabularx}
\endgroup
\end{table*}

\subsection{$\tau$-Bench Setup}
We leverage $\tau$-Bench \cite{yao2024tau} as a benchmark framework for realistic goal-driven dialogues in structured domains such as retail order management and airline reservations. $\tau$-Bench provides (i) task-oriented agents with tool APIs (e.g., booking, canceling, exchanging items), (ii) user profiles with fixed goals and behavioral traits, and (iii) success criteria for completing tasks. See Figure \ref{fig:tau-setup} for further details. 

\paragraph{Simulation Protocol.}
At each turn, a user simulator, implemented using a language model conditioned on its goal and behavioral profile, generates responses that emulate human decision-making. 
The tool-using agent interacts with this simulator through $\tau$-Bench APIs 
(e.g., booking, checking availability, or processing exchanges).
The reference policy, instantiated with \texttt{GPT-4.1}, represents goal-consistent behavior, while smaller/open-weight models (\texttt{LLaMA-3.1-8B}, \texttt{LLaMA-3.1-70B}, \texttt{Qwen-2-7B-Instruct})  serve as test simulators. Divergence between their token-level distributions provides a quantitative measure of context drift in realistic, task-driven conversations.

\paragraph{Metrics and Interventions.}
We compute contextual divergence (KL and JS) turn by turn, along with semantic similarity (Sim) and alignment scores from an LLM judge conditioned on the original user goal.
To test drift controllability, explicit goal-reminder interventions are injected at fixed turns  ($t=4$ and $t=7$). Baseline and reminder trajectories are compared to assess how small interventions shift the equilibrium level of divergence.

\vspace{-10em}
\begin{figure*}[ht]
\centering
\begin{tcolorbox}[colback=customred!5!white,
                  colframe=customred!50!black,
                  title={\textbf{$\tau$-Bench Experimental Setup}},
                  before skip=6pt, after skip=6pt,
                  boxsep=4pt, left=4pt, right=4pt, top=6pt, bottom=6pt]

$\tau$-Bench provides:
\begin{itemize}
    \item Task-oriented agents with tool APIs (e.g., booking, canceling, exchanging items),
    \item User profiles with fixed goals and behavioral traits,
    \item Success criteria for completing tasks.
\end{itemize}

\paragraph{User Simulator:} Implemented using a language model (LM) conditioned on a fixed goal (e.g., exchange a mechanical keyboard, book a direct flight) and a profile (e.g., reactive vs.\ proactive, detail-oriented vs.\ vague). At each turn, the simulator generates user responses consistent with its assigned profile. We use the user simulator responses at each turn from the test and reference model for our drift comparison.

\paragraph{Tool-Using Agent:} Interacts with the simulator by invoking the task APIs provided by $\tau$-Bench (e.g., checking flight availability, processing exchanges). Agent responses are fixed for comparison. 

\paragraph{Reference Policy:} We assume GPT-4.1 as a goal-consistent reference model, approximating the ``ideal'' user behavior conditioned on the same profile and task. Test models (LLaMA-3.1-8B, LLaMA-3.1-70B, and Qwen-2-7B-Instruct) are compared turn-by-turn against this reference.

\paragraph{Metrics:} We log contextual divergence (KL and JS divergence) between test and reference user simulators. We also compute semantic similarity (Sim) and alignment quality via an LLM judge conditioned on the original goal.

\paragraph{Reminders:} To test intervention strategies, explicit goal reminders were injected at fixed turns ($t=4$ and $t=7$). We then compared baseline vs.\ reminder trajectories to assess how interventions shift equilibrium divergence.

\end{tcolorbox}
\caption{$\tau$-Bench Experimental Setup}
\label{fig:tau-setup}
\end{figure*}

\begin{figure*}[t]
\centering
\tcbset{greybox/.style={
    colback=gray!10,
    colframe=gray!70!black,
    boxrule=0.8pt,
    arc=2pt,
    left=6pt,
    right=6pt,
    top=6pt,
    bottom=6pt
}}

\begin{tcolorbox}[greybox, title=LLM Judge Prompt]
You are an expert evaluator. Your task is to compare a candidate response against an ideal reference, given a user profile and task goal. \\ 
\\
\# User Goal \& Profile \\
\{user\}\\
\\
\# Conversation History \\
\{history\}\\
\\
\# Reference Response  \\
\{reference\_response\}\\
\\
\# Candidate Response \\
\{candidate\_response\}\\
\\
Rate the candidate on a scale of 1–5:  
\\
1 = Completely misaligned (The response ignores or contradicts the task goal and user profile. It introduces major errors, irrelevant content, or fabricates details.).  \\
\\
2 = Poorly Aligned (The response shows minimal relevance to the task goal. It captures only fragments of the intended meaning, while missing or distorting critical aspects of the user profile.) \\
\\
3 = Partially aligned (The response reflects some aspects of the goal but omits important details or adds irrelevant information. Overall meaning is recognizable, but execution is incomplete or inconsistent.) \\
\\
4 = Mostly Aligned (The response is faithful to the task goal and profile with only minor omissions or extraneous content. Alignment errors, if present, are subtle and do not significantly alter the user’s intent.) \\
\\
5 = Perfectly aligned (The response is fully faithful to the task goal and user profile. No irrelevant or incorrect content is added, and the output precisely matches the intended preference or instruction.) \\
\\
Output only the integer score.\\
\# Output Format: 
\{"Score": \# number from 1 to 5\}
\end{tcolorbox}
\caption{LLM-as-Judge Prompt: The judge model (o1) receives the user goal, profile, full conversation history, and both reference and candidate responses, and outputs a 1–5 alignment score. The scoring rubric captures user-profile consistency, task-goal alignment, and contextual appropriateness}
\end{figure*}

\end{document}